\documentclass{article}

\usepackage{microtype}
\usepackage{graphicx}
\usepackage{subfigure}
\usepackage{booktabs} 

\usepackage{hyperref}

\usepackage[accepted]{icml2025}

\usepackage{amsmath}
\usepackage{amssymb}
\usepackage{mathtools}
\usepackage{amsthm}

\usepackage[utf8]{inputenc} 
\usepackage[T1]{fontenc}    
\usepackage{booktabs}       
\usepackage{amsfonts}       
\usepackage{nicefrac}       
\usepackage{microtype}      
\usepackage{xcolor}         

\usepackage{enumitem}
\usepackage{dsfont}
\usepackage{amsthm}
\usepackage{amssymb}

\usepackage[noend]{algpseudocode}
\usepackage{mathtools}
\usepackage{makecell}
\usepackage{multirow}
\usepackage{subcaption}
\usepackage{color}
\usepackage{colortbl}
\usepackage{xspace}

\usepackage[capitalize,noabbrev]{cleveref}

\theoremstyle{plain}

\theoremstyle{definition}

\theoremstyle{remark}

\usepackage[textsize=tiny]{todonotes}

\icmltitlerunning{CDR-Agent}

\begin{document}

\twocolumn[
\icmltitle{CDR-Agent: Intelligent Selection and Execution of Clinical Decision Rules Using Large Language Model Agents}



\icmlsetsymbol{equal}{*}

\begin{icmlauthorlist}
\vspace{-0.15in}
\icmlauthor{Zhen Xiang}{equal,uga}
\icmlauthor{Aliyah R. Hsu}{equal,ucb}
\icmlauthor{Austin V. Zane}{ucb}
\icmlauthor{Aaron E. Kornblith}{ucsf}
\icmlauthor{Margaret J. Lin-Martore}{ucsf}
\icmlauthor{Jasmanpreet C. Kaur}{ucsf}
\icmlauthor{Vasuda M. Dokiparthi}{ucsf}
\icmlauthor{Bo Li}{uchi}
\icmlauthor{Bin Yu}{ucb}
\end{icmlauthorlist}

\icmlaffiliation{uga}{University of Georgia}
\icmlaffiliation{ucb}{University of California, Berkeley}
\icmlaffiliation{ucsf}{University of California, San Francisco}
\icmlaffiliation{uchi}{University of Chicago}

\icmlcorrespondingauthor{Zhen Xiang}{zxiangaa@uga.edu}
\icmlcorrespondingauthor{Bin Yu}{binyu@berkeley.edu}

\icmlkeywords{Machine Learning, ICML}

\vskip 0.3in
]



\printAffiliationsAndNotice{\icmlEqualContribution} 

\begin{abstract}
Clinical decision-making in emergency departments (EDs) is complex, rapid, and high-stakes.
Clinical Decision Rules (CDRs) are standardized evidence-based tools that combine signs, symptoms, and clinical variables into decision trees to make consistent and accurate diagnoses.
CDR usage is often hindered by the clinician's cognitive load, limiting their ability to quickly recall and apply the appropriate rules.
Moreover, current AI systems for clinical support struggle to consistently follow medical knowledge, adapt to safety policies, and coordinate complex decision-making.
We introduce CDR-Agent, an LLM-based system that autonomously identifies and applies the appropriate CDRs from unstructured clinical notes while maintaining efficiency, robustness, and structured knowledge access.
To validate CDR-Agent, we curated two novel ED datasets: synthetic and CDR-Bench, although CDR-Agent is applicable to non ED clinics.
CDR-Agent improves CDR selection accuracy by 56.3\% on the synthetic dataset and 8.7\% on CDR-Bench compared with a standalone LLM baseline, while also reducing computational overhead.
Using these datasets, we demonstrated that CDR-Agent not only selects relevant CDRs efficiently, but makes cautious yet effective imaging decisions by minimizing unnecessary interventions while successfully identifying most positively diagnosed cases, outperforming traditional LLM prompting approaches.

\vspace{0.15in}
\end{abstract}

\section{Introduction}\label{sec:introduction}

Clinical Decision Rules (CDRs) are standardized tools designed to assist clinicians by combining signs, symptoms, and clinical features into decision trees, enabling accurate and consistent, evidence-based bedside decisions \cite{Pines2023}.
Currently, over 700 CDRs span nearly all medical specialties, covering diverse clinical scenarios  \cite{Obra2025, Heerink2023}.
These rules utilize patient data to generate composite scores or assessments that stratify patient risk for disease onset, progression, or clinical outcomes, helping clinicians deliver expert-level care regardless of their level of experience \cite{Holmes2014, Chan2020}. 

However, across healthcare settings, clinicians face increasing time pressure and cognitive burden when evaluating patients.
Whether in outpatient clinics, inpatient wards, or telemedicine encounters, clinicians must rapidly assess complex presentations with limited time and incomplete information.
In these environments, tools that can surface the right decision aid at the right time can help mitigate errors and democratize care quality. This challenge is even more pronounced in high-stakes trauma care, where rapid and accurate decisions are essential to avoid the harms associated with missed critical injuries or unnecessary imaging and interventions \cite{Neumann2024}.
Although trauma care has been regionalized to concentrate expertise, injured patients often initially present to emergency departments (EDs) without trauma specialization, further emphasizing the need for universally applicable decision-making aids like CDRs \cite{Deshormes2024, Perry2006}.

Despite the importance of CDRs, clinicians often face significant challenges recalling and applying them appropriately due to the sheer number of rules tailored to specific clinical conditions, organ systems, and patient populations \cite{Kharel2023}. This challenge is not confined to emergency care -- clinicians across healthcare settings routinely make rapid decisions under time constraints, often with incomplete data. Whether in prehospital, outpatient, or inpatient medicine, clinicians must balance efficiency with accuracy, making real-time access to relevant CDRs critically important.

Recent advancements have highlighted the integration of Large Language Models (LLMs) into Clinical Decision Support Systems (CDSS), significantly enhancing clinical decision-making processes. LLMs augmented with external medical knowledge, such as literature databases and clinical guidelines, demonstrate superior performance compared to standalone models in various clinical tasks, including COVID-19 outpatient care, medication prescriptions, and diagnostic accuracy \cite{Wang2024, Ong2024, Lammert2024}. Additionally, explanations generated by LLMs based on patient notes have been shown to improve clinician agreement rates, emphasizing their potential utility in clinical contexts \cite{Umerenkov2023}. Nevertheless, existing approaches often suffer from limited generalizability due to their application within narrow, specialized healthcare settings \cite{Panicker2023}.
Particularly relevant to our work is the study by Zakka et al., which introduced an LLM-based agent with external internet access and an extensive database of clinical calculators derived from MDCalc, coupled with a ClinicalQA dataset, achieving substantial performance improvements over plain LLMs \cite{Zakka2024}.

In this paper, we focus on trauma care as a case study -- not because it is exclusive to the ED, but because it uniquely spans multiple disciplines and involves numerous CDRs across different organ systems and clinical specialities (e.g. trauma surgery, orthopedics, critical care, radiology, etc.).
This allows us to explore whether an LLM-based agent can support holistic clinical reasoning by recognizing when and how to apply a diverse set of CDRs within a single patient scenario, which is a challenge that remains largely unmet in current AI systems.

Compared to existing research focused on automated diagnosis \cite{abbasian2024conversational, shi2024ehragent, mcduff2023differential, tu2024diagnosticai, LEVRA2025113}, knowledge-intensive medical question answering \cite{SinghalKaran2023Llme, saab2024medgemini, Wang2023}, clinical data analysis \cite{Xiao2025.06.01.25328537}, and medication management \cite{Ong2024}, emergency department decision-making uniquely prioritizes accuracy and efficiency amidst challenges posed by incomplete patient information. Prior studies that augment LLMs with external medical knowledge typically employ retrieval-augmented generation (RAG) methods, which select relevant resources to contextualize the LLM's responses. Although these methods improve accuracy and efficiency relative to standalone models, this alone is insufficient in emergency department scenarios. Existing RAG approaches commonly lack transparency regarding how the LLM integrates retrieved information or verifies the appropriateness of selected resources, limiting interpretability. This absence of clear rationale undermines clinical trust, as emergency clinicians rely heavily on transparent, interpretable decision-making to manage rapid, accurate patient care amidst incomplete and evolving clinical information.

As a remedy, we propose CDR-Agent, an LLM-based system designed to autonomously select and apply the most suitable clinical decision rules (CDRs) based on given clinical notes in real ED situations.
CDR-Agent's operation involves three primary steps, closely mirroring the decision-making process of clinicians. First, it identifies relevant CDRs by measuring semantic similarity between an input clinical note and available CDR descriptions. To maintain computational efficiency, this step leverages an embedding model without invoking LLM queries. The system then extracts specific indicator values from the clinical note, which serve as input variables for the selected CDRs followed by a verification of exclusion criteria to determine the CDRs valid to proceed. If certain necessary information is missing, clinicians may either collect the information or infer it using their clinical expertise. Finally, CDR-Agent precisely executes the selected CDRs by running the Python script corresponding to each of them to generate decision outcomes.

Due to the lack of public benchmarks for validating our CDR-Agent, we evaluate its effectiveness and efficiency on two curated ED datasets: CDR-Bench and a synthetic dataset derived from Pediatric Emergency Care Applied Research Network (PECARN) \cite{Holmes2013LowRiskAbdominalInjury, Kuppermann2009LowRiskBrainInjury}.
The synthetic dataset provides a controlled evaluation setting, where each note is generated from PECARN tabular data and contains a single CDR ground truth label with no missing values. This ensures a structured assessment of model performance in unambiguous clinical scenarios.
CDR-Bench consists of real-world clinical notes from various public datasets, annotated with clinician-labeled CDRs. These notes exhibit diverse writing styles, contain noise, and often have missing values, making CDR-Bench a challenging real-world robustness test. CDR-Bench features an adaptive number of CDRs per note, better reflecting real-world clinical decision-making complexity.
Together, these datasets offer a comprehensive evaluation of CDR-Agent's ability to interpret clinical narratives and apply trauma-related CDRs accurately. Both datasets are released with code and detailed documentation, ensuring reproducibility and facilitating further research in LLM-driven clinical decision support.

We evaluate CDR-Agent on the two datasets against a baseline approach that directly queries the same LLM for CDR execution results using the clinical notes and a list of available CDRs.
Using these datasets, we demonstrated that CDR-Agent not only selects relevant CDRs efficiently, but makes cautious yet effective imaging decisions by minimizing unnecessary interventions while successfully identifying most positively diagnosed cases, outperforming traditional LLM prompting approaches.
Moreover, the entire procedure of CDR selection and execution costs 1.22 and 1.16 seconds on the synthetic data and CDR-Bench, respectively, much faster than the baseline with 4.12 and 8.7 seconds on the two datasets.

In summary, we propose CDR-Agent, the first automated, LLM-based system that assists with expert-level decision-making using CDRs in ED scenarios.
We also created two datasets to facilitate research on ED decision-making involving CDR.
Finally, our in-depth case studies underscore the potential to foster collaboration between clinicians and technicians in this field.
Code for our work can be found at: \url{https://github.com/zhenxianglance/medagent-cdr-agent}

\begin{figure*}[t]
  \centering
    \includegraphics[width=\textwidth]{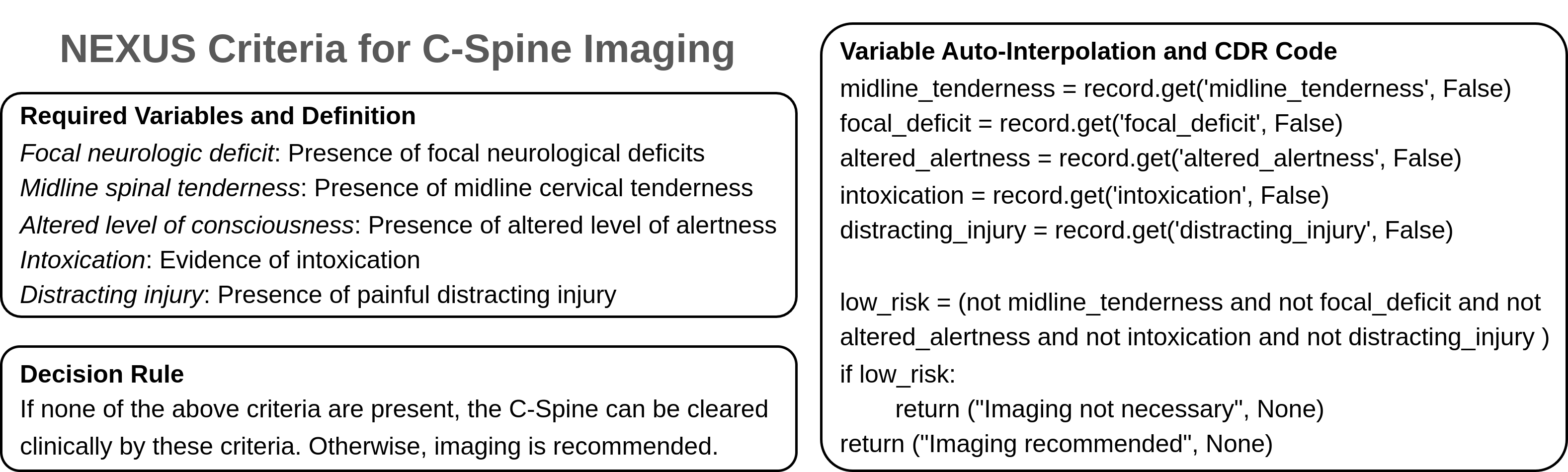}
 \caption{An example CDR for C-spine imaging. Top left: the variables/indicators required by the CDR and their definitions. Bottom left: the rule deciding whether the patient requires imaging. Right: the Python script for the CDR with automated imputation of missing variables.}
 \label{fig:cdr_illustration}
\end{figure*}

\section{Problem Description}\label{sec:problem}

To set up the task, we consider a clinical note $\boldsymbol{x}$ that describes patient information such as demographics, chief complaints, and physical exam results, and a set of predefined CDRs $\mathcal{C}=\{c_1, \cdots, c_N\}$.
Each CDR $c_i$ is a decision tree function that maps from a specific variable space $\mathcal{V}_{c_i}$ to a set of predefined outcomes.
Here, the variables are the indicators in the clinical notes required by the CDR to arrive at a decision outcome.
For example, the NEXUS criteria for C-spine imaging require five binary indicators: presence of focal neurologic deficit, midline spinal tenderness, altered level of consciousness, intoxication, and distracting injury, respectively, to determine an outcome that can either be ``imaging recommended'' (if any of the indicators present) or ``imaging not necessary'' (if no indicator is found), as shown in Figure \ref{fig:cdr_illustration}.
Our design objective is to build an automated system that, for any input clinical note $\boldsymbol{x}$, identifies the most appropriate CDRs from $\mathcal{C}$, if there are any, and then execute the identified CDRs to get a set of final decisions.

\section{Methods}\label{sec:methods}

In this section, we introduce our CDR-Agent framework proposed to address the problem described above, focusing on the design details and underlying rationales.

\subsection{Overview}
Our system design takes into account real-world clinical scenarios with the following challenges.
First, clinical notes often vary significantly in writing style and terms, making it difficult to accurately identify relevant indicators or variables for CDRs through simple word matching.
Second, 
some critical information may be incomplete, particularly in cases involving unconscious, disabled, or pediatric patients.
Third, the execution of CDRs is prone to errors, both by humans, due to the complexity of the rules, and by automated systems such as LLMs, which may misinterpret textual rule descriptions and generate incorrect decisions.

\begin{figure*}[t]
  \centering
    \includegraphics[width=1.0\textwidth]{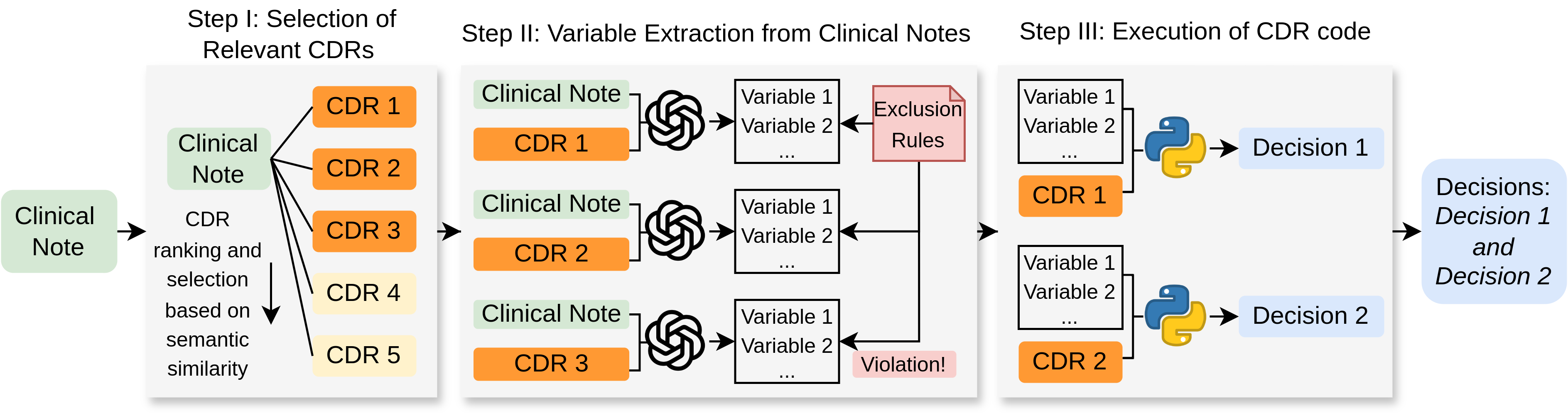}
 \caption{Illustration of the three-step workflow of CDR-Agent.
 For any input clinical note, CDR-Agent first selects a number of relevant CDRs that have high semantic similarity to the clinical note. Variables required by each selected CDR are then extracted from the clinical note using an LLM, with a set of exclusion rules applied to filter invalid CDRs. Finally, CDR-Agent executes the Python code for each valid CDR for decisions.}
 \label{fig:figure1}
\end{figure*}

CDR-Agent follows a three-step workflow designed to address these challenges, and we descibe the deatils of each step in the following paragraphs. The complete pipeline of CDR-Agent is shown in Figure \ref{fig:figure1}.

\subsection{CDR Selection Step}
The goal of the first step is to identify the most relevant CDRs for a given clinical note.
We use \textit{semantic similarity} as a proxy to measure the relevance between the clinical note and CDRs.
To measure semantic similarity, we first converted the CDR functions into textual descriptions
using GPT-4o followed by clinician
inspection to ensure the correctness, and used a pretrained word embedding model, text-embedding-ada-002 by OpenAI, to map both the clinical notes and the textual description of CDRs
into a shared embedding vector space.

Formally, for each CDR $c_i$, we compute an embedding\footnote{The actual computation is to map each token in the text to an embedding vector and then take the average embedding vector over the entire text.} vector $E(\boldsymbol{x})$ for the clinical note $\boldsymbol{x}$ and an embedding vector $E(\boldsymbol{t}_i)$ for the textual description $\boldsymbol{t}_i$ of the CDR, respectively.
Here, the embedding model is trained by OpenAI
on a large corpus of text data, including medical literature, ensuring that semantically similar content is mapped to vectors with higher cosine similarity.
A straightforward approach to select CDRs would be to compute the cosine similarity $s^{(i)}=\text{cosim}(E(\boldsymbol{x}), E(\boldsymbol{t_i}))$ for each CDR and select the top-$k$ CDRs with the highest similarity scores for a predefined $k$. However, setting a value of $k$ or a naive threshold on similarity scores for CDR selection is unrealistic and challenging in practice because of the variability in similarity score ranges across different clinical notes.

As a solution, we propose an anomaly detection approach to identify for each clinical note the CDRs with abnormally large similarity scores for selection in a principled way.
Specifically, for each clinical note, we fit a Gaussian distribution using the computed similarity scores (the choice of Gaussian is validated in Section ``Deisgn Choices and Further Improvements'').
A CDR $c_i$ is selected as an anomaly if $p(s^{(i)}|\mu, \sigma^2)>\alpha$, where $\mu$ and $\sigma$ are the estimated mean and variance for the Gaussian distribution, and $\alpha$ is a predefined significance level.
Drawing inspiration from hypothesis testing, we set $\alpha=0.05$ in all our experiments, which achieves an effective balance between minimizing false positives and enabling the detection of real outliers.
If no CDR is identified as an anomaly, it indicates that none of the available CDRs are applicable to the given clinical scenario.

In addition to the above-mentioned base design, we will demonstrate in Section ``Deisgn Choices and Further Improvements'' that multiple features can be added to the framework to further improve the CDR selection accuracy.

\subsection{Variable Extraction Step}
\begin{figure*}[t]
  \centering
    \includegraphics[width=\textwidth]{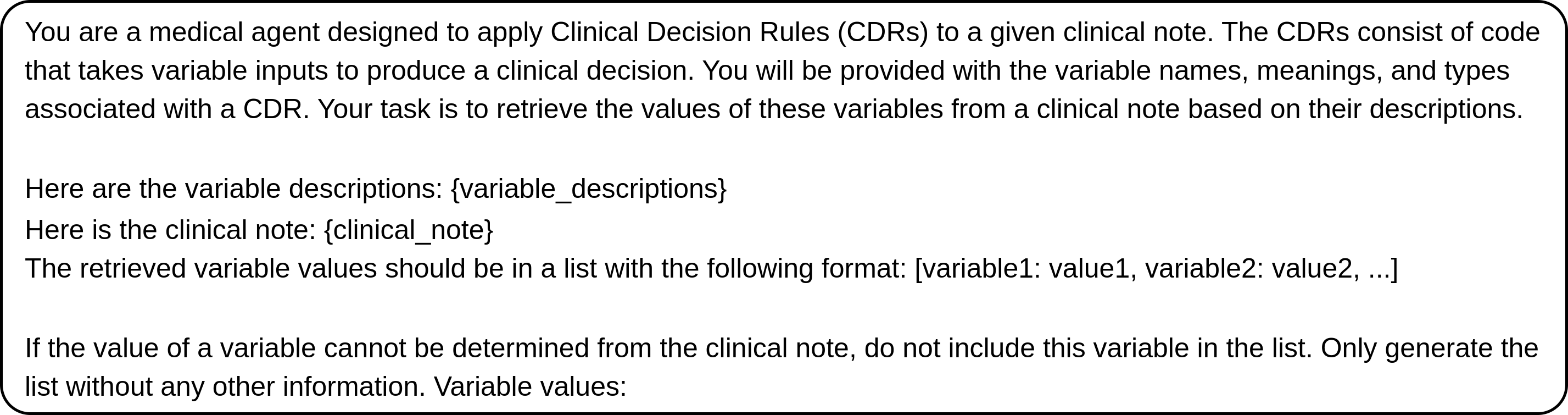}
 \caption{Prompt to LLM for variable extraction from a given clinical note. The prompt includes variables required by the selected CDR and their definitions, and the formatting requirements for the extracted variable values.}
 \label{fig:var_extraction}
\end{figure*}

CDR-Agent uses an LLM to extract the variables required by each selected CDR from the given clinical note, with a prompt shown in Figure \ref{fig:var_extraction}.
The key to this prompt design is the clear specification of variable \textit{definitions} and \textit{formatting}.
For each selected CDR, the prompt includes the definition of each variable (see top-left of Figure \ref{fig:cdr_illustration} for example) to help the LLM accurately interpret its meaning while searching for the corresponding values from the clinical note.
Additionally, the extracted variables are structured in a list format, where each variable name is followed by its corresponding extracted value.

To handle potential missing values, we include only the determined variables in the output list.
In practice, absent variables can be filled in by consulting clinicians for additional information collection from patients.
However, in our experiments with \textit{automated} evaluation settings, this feature (which requires human participation) is not incorporated.
Instead, we adopt a ``negative'' imputation approach, where missing variables are assigned default values that do not trigger positive clinical decisions, such as imaging recommendations.
This practice prevents the LLM from arbitrarily assigning values that could lead to false positives.
For future work, we will explore alternative imputation strategies that leverage external knowledge, such as population-level measurements, to enhance accuracy of CDR execution.
After extracting the variables, a set of exclusion rules, according to the inclusion / exclusion criteria for each CDR, is applied to filter out invalid CDRs, such as excluding adult patients for PECARN CDRs.

\subsection{CDR Execution Step}
CDR-Agent executes each valid CDR by running a predefined Python script (see the right of Figure \ref{fig:figure1} for an example Python script), using the variable values extracted from the clinical note in step 2.
Before execution, the extracted variable values are converted to their predefined data types, including boolean, integer, float, and string, ensuring consistency and correctness.
Formally, the decision outcome for a selected CDR $c_i$ is obtained as $y_i=f_{c_i}(\{v_1^{(i)}, \cdots, v_L^{(i)}\})$, where $\{v_1^{(i)}, \cdots, v_L^{(i)}\}$ are the variable values after formatting and $f_{c_i}$ is the CDR function.
Note that any execution failure will trigger an error message, allowing the agent to prompt manual intervention to verify the information processed in earlier steps.
This design ensures that only outputs based on accurate variable extraction are produced, minimizing potential errors caused by LLM hallucinations.
Once all valid CDRs have been executed, their decisions are aggregated into a final list as the next-step management for the patient.

\section{Dataset Construction}\label{sec:datasets}

As the first study to explore the automation of CDR selection and execution in ED clinical scenarios, we develop two datasets to evaluate our framework and serve as a benchmark for future research. Collectively comprising 544 patient notes, our dataset significantly surpasses prior related datasets by an order of magnitude.

\begin{figure*}[t]
    \centering
    \begin{minipage}{0.58\linewidth}
        \centering
        \includegraphics[width=\linewidth]{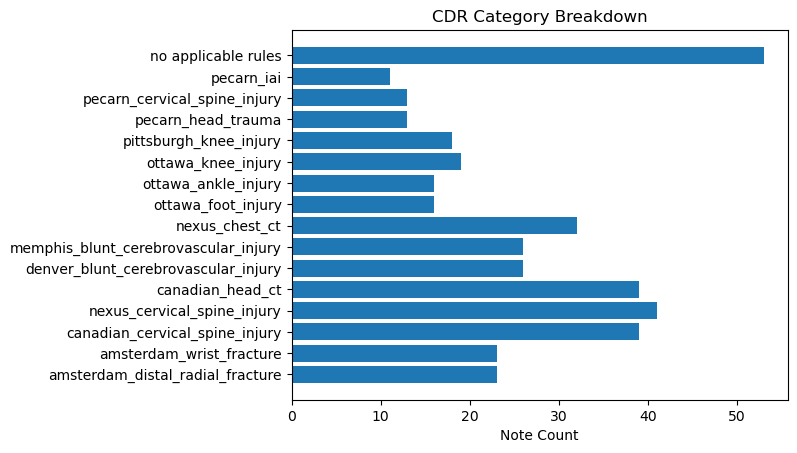}
    \end{minipage}%
    ~ 
    \begin{minipage}{0.4\linewidth}
        \centering
        \includegraphics[width=\linewidth]{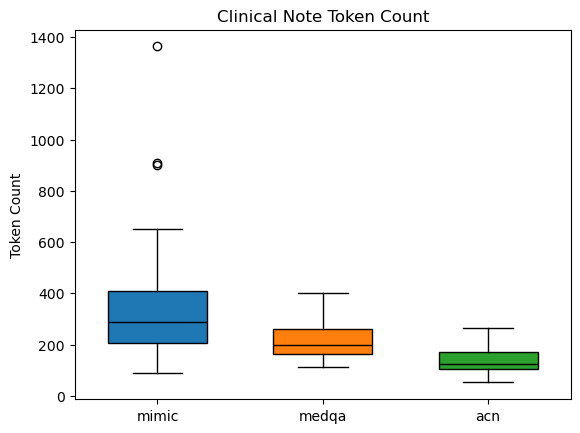}
    \end{minipage}
    \caption{(Left) A detailed breakdown of CDR label composition in CDR-Bench. Approximately 36.8\% of the notes have no applicable CDRs.
    (Right) Token length variations across different data sources in CDR-Bench. MIMIC-IV notes are significantly longer than those from MedQA and ACN, often containing more noise and distracting information. This highlights both the diversity captured in CDR-Bench and the challenges it presents for CDR selection.}
    \label{fig:cdr_bench_stats}
\end{figure*}

\paragraph{CDR Database}
Based on clinician recommendations, we selected 15 trauma-related CDRs from peer-reviewed publications to form our CDR database (see Fig. \ref{fig:cdr_bench_stats} for a complete list). Each rule was manually transcribed from its original published format into a structured Python function, enabling automated execution within our system. To facilitate accurate interpretation and execution, we also provided detailed textual descriptions for each input feature and for the rule's clinical purpose. This standardized approach ensures that models evaluated on this dataset have sufficient context to correctly select and apply relevant rules. The resulting CDR database was utilized for our experiments on both CDR-Bench and the synthetic dataset, demonstrating the generalizability of our method.

\paragraph{Synthetic Dataset}
PECARN provides Clinical Decision Rules (CDRs) along with corresponding tabular datasets for pediatric traumatic brain injuries (TBI) and pediatric intra-abdominal injuries (IAI). To systematically evaluate our system under controlled conditions, we generated a synthetic dataset by converting tabular patient data from PECARN into realistic free-form clinical narratives. Specifically, we randomly selected a total of 400 patients (200 TBI and 200 IAI), of which 20\% required medical intervention, ensuring representation of clinically significant scenarios. We first use templates to convert binary patient features into structured sentences, then refine these into coherent, natural-sounding clinical notes using GPT-4o. The resulting clinical narratives averaged approximately 98 tokens in length.

\paragraph{CDR-Bench}
CDR-Bench is constructed using real-world clinical notes from three well-established public datasets: MIMIC-IV \cite{johnson2023mimic}, MedQA \cite{jin2020disease-medqa}, and Augmented Clinical Notes (ACN) \cite{augmented_clinical_notes-acn}, each featuring diverse writing styles. To ensure a balanced dataset with both CDR-applicable and non-applicable cases, we filter for notes containing trauma-related keywords (e.g., fracture, injury, trauma). This step prevents an overwhelming number of notes without relevant CDRs, given the large dataset sizes.

We randomly selected 50 trauma-related notes from each source for annotation by four emergency medicine clinicians. Initially, each sample was independently labeled by two clinicians. In cases of disagreement (with an average inter-annotator agreement of ~80\%), a third, more senior clinician reviewed the discrepancies and provided a final adjudication.
Recognizing that clinical decision-making often lacks a single ground truth, we retain all label sets from the annotators. During evaluation, we measure the model's maximum accuracy across these label sets, meaning CDR-Agent is considered correct if it aligns with any clinician’s judgment. This approach reflects the inherent variability in medical decision-making while ensuring a robust evaluation of CDR selection performance.

After notes with incomplete information to label were removed, following the clinicians' feedback, the final CDR-Bench consists of 144 samples (MIMIC-IV: 47, MedQA: 47, ACN: 50).
On average, each note contains 233.9 tokens. Approximately 36.8\% of the notes have no applicable CDRs, ensuring a mix of relevant and irrelevant cases for robustness testing. Among the labeled samples, the average number of CDRs per note is 2.83, highlighting the multi-label nature of clinical decision-making.
These facts together make CDR-Bench a challenging dataset compared to the synthetic dataset.
See Figure \ref{fig:cdr_bench_stats} for a detailed breakdown of CDR label composition and token length variations across different data sources in CDR-Bench.

\begin{table*}[t!]
\caption{Evaluation results for CDR-Agent on the synthetic data and CDR-Bench, compared with the baseline approach purely based on LLM querying.
Evaluation metrics including the exact match (EA) accuracy, F1-score, and time cost ($T_{\text{sel}}$) for CDR selection, and the sensitivity, specificity, and time cost ($T_{\text{exe}}$) for CDR execution. The total time costs ($T_{\text{tot}}$) for both methods are also reported.
All time costs are in \textit{second}.
``n.a.'' means ``not applicable''.
}\label{tab:main_results}
\small
\begin{center}
\begin{tabular}{|l|l|l|l|l|l|l|l|l|}
\hline
\multicolumn{2}{|l|}{} & \multicolumn{3}{c|}{CDR Selection} & \multicolumn{3}{c|}{CDR Execution} & \\
\cline{3-9}
\multicolumn{2}{|l|}{} & EA Accuracy & F1-Score & $T_{\text{sel}}$ & Sensitivity & Specificity & $T_{\text{exe}}$ & $T_{\text{tot}}$\\
\hline
\multirow{2}{*}{Synthetic Data} & Baseline & 0.420 & 0.735 & n.a. & 0.818 & 0.708 & n.a. & 4.12\\
\cline{2-9}
& \cellcolor{gray!15}CDR-Agent & \cellcolor{gray!15}0.983 & \cellcolor{gray!15}0.994 & \cellcolor{gray!15}0.44 & \cellcolor{gray!15}0.983 & \cellcolor{gray!15}0.687 & \cellcolor{gray!15}0.78 & \cellcolor{gray!15}1.22\\
\hline
\multirow{2}{*}{CDR-Bench} & Baseline & 0.426 & 0.608 & n.a. & 0.936 & 0.652 & n.a. & 8.70\\
\cline{2-9}
& \cellcolor{gray!15}CDR-Agent & \cellcolor{gray!15}0.513 & \cellcolor{gray!15}0.592 & \cellcolor{gray!15}0.52 & \cellcolor{gray!15}0.683 & \cellcolor{gray!15}0.983 & \cellcolor{gray!15}0.64 & \cellcolor{gray!15}1.16\\
\hline
\end{tabular}
\end{center}
\end{table*}

\section{Experimental Setup}\label{sec:setup}

We aim to evaluate the proposed CDR-Agent with a focus on two research questions:
First, can CDR-Agent \textit{accurately} and \textit{efficiently} select all relevant CDRs for a given clinical note?
Second, with the incorporation of LLMs, can CDR-Agent \textit{accurately} and \textit{efficiently} extract the required variables from the complex clinical note and execute the CDRs reliably?
Note again that efficiency here refers to the computational time cost of the system. 

To answer these questions, we use two sets of metrics to assess CDR selection and execution outcomes, respectively.
Note that the CDR labeling for a clinical note can result in a single CDR, a set of CDRs, or ``no applicable CDR''.
We evaluate CDR selection using three metrics: 1) \textbf{exact match (EA) accuracy}, which measures the proportion of clinical notes where the selected CDR(s) exactly match the labeled CDR(s); 2) \textbf{F1-score}, which jointly assesses the precision and recall for CDR selection\footnote{Here, we treat CDR selection for each clinical note as a binary classification problem over the set of all candidate CDRs, including ``no applicable CDR''. The actual positives are the applicable CDRs (or ``no applicable CDR''), and a true positive is counted when a selected CDR matches the ground truth.}; and 3) \textbf{selection time}, which quantifies the computational cost of CDR selection in seconds.
For CDR execution, we focus on sensitivity and specificity: 1) \textbf{sensitivity}, which measures the sensitivity of outcome recommendation across all \emph{correctly selected} CDRs; 2) \textbf{specificity,} which measures the specificity across all \emph{correctly selected} CDRs; and 3) \textbf{execution time}, which measures the average time cost for each CDR execution, including both variable extraction and execution of Python scripts of CDRs.

To illustrate the benefits of our design over conventional LLM querying, we compare CDR-Agent with a baseline approach where an LLM is directly queried using the clinical note alongside all the CDR information to determine the final CDR execution outcomes.
We report the main results in Table \ref{tab:main_results}, where we used one of the state-of-the-art LLMs, GPT-4o, as the core LLM for both approaches.
Note that for some other LLM choices, such as GPT-4, the baseline approach easily encounters maximum token limitations due to the extensive prompt length required by the inclusion of all CDRs.
This issue will be further amplified as the number of CDRs increases.
However, our CDR-Agent is not affected by this issue as it leverages an embedding model to generate embedding vectors to compute similarity scores for CDR selection, without the need of squeezing all CDR information in-context in a query.
In addition to the main results, we also evaluate CDR-Agent, particularly its CDR execution performance on other LLM choices (as model choices don't affect CDR selection performance), including GPT-4 and GPT-4o-mini, and report the results in Table \ref{tab:model_choice}.




\section{Results and Findings}\label{sec:results}

\paragraph{Main Evaluation Results} The comparison between CDR-Agent and the baseline approach for CDR selection and execution is presented in Table \ref{tab:main_results}.
For CDR selection, on the synthetic dataset, CDR-Agent achieves an exact match accuracy of 0.983 and an F1-score of 0.994, both significantly outperforming the baseline. On CDR-Bench, CDR-Agent achieves a superior exact match accuracy of 0.513 compared to 0.426 for the baseline, while maintaining a comparable F1-score. Additionally, CDR-Agent is highly efficient, requiring only 1.19 seconds on average to complete the entire procedure, compared to 6.41 seconds for the baseline.

For CDR execution, it is important to note that the ground truth ``outcome" labels used to compute sensitivity and specificity differ in nature between the two datasets. In the synthetic dataset (sourced from PECARN tabular data), outcome labels indicate whether a patient was actually diagnosed with TBI or IAI, regardless of whether they received an imaging intervention. However, such diagnosis outcome labels were not available in the clinical notes in our collected CDR-Bench. As a proxy, we use GPT-4o to infer whether a patient received the corresponding imaging intervention recommended by a CDR.

We suspect that this difference in outcome label definitions explains the variation in CDR-Agent’s predictive sensitivity and specificity across the two datasets. On CDR-Bench, CDR-Agent achieves high specificity (0.983) and moderate sensitivity (0.683), suggesting a more conservative behavior compared to a vanilla LLM. This conservatism is often due to insufficient information in clinical notes to confidently recommend imaging. It should not be misunderstood as a failure to identify positive diagnoses.
In contrast, on the synthetic dataset, where diagnosis labels are directly available, CDR-Agent achieves high sensitivity (0.983) and decent specificity (0.687), indicating strong performance in identifying true positive diagnoses. This high sensitivity is particularly desirable in medical settings, where the primary goal is to avoid missing serious or life-threatening conditions.

Overall, CDR-Agent's performance in CDR execution across both datasets highlights its ability to make cautious yet effective imaging decisions, minimizing unnecessary imaging while still capturing most positively diagnosed cases, outperforming the baseline in both respects.
Moreover, CDR-Agent’s performance on CDR-Bench improves further when using GPT-4 as the core LLM (Table \ref{tab:model_choice}), increasing sensitivity to 0.786 and specificity to 1.0, with only a one-second increase in time cost. In contrast, the baseline does not benefit from using GPT-4 due to its limited scalability with the number of CDRs, constrained by model input limits. In terms of computation time, CDR-Agent is over seven times faster than the baseline, demonstrating its efficiency.

In summary, CDR-Agent matches or outperforms the baseline in both CDR selection and execution, while achieving significantly lower computational cost, underscoring its suitability for real-time ED decision-making.

\paragraph{Influence of LLM Choices} Table \ref{tab:model_choice} presents the performance of CDR-Agent in CDR execution using different core LLMs.
The results for CDR selection are omitted here, as this step does not involve querying an LLM.
On the synthetic dataset, CDR-Agent achieves comparable sensitivity and specificity when using GPT-4o and GPT-4, with GPT-4 exhibiting slightly more conservative behavior.
However, CDR-Agent with GPT-4o demonstrates significantly better efficiency, requiring only about one-quarter of the execution time compared to GPT-4.
GPT-4o-mini achieves sensitivity comparable to GPT-4o but with noticeably lower specificity and increased execution time.
On CDR-Bench, both the sensitivity and specificity are clearly improved when changing the core LLM from GPT-4o or GPT-4o-mini to GPT-4.
These findings suggest that the performance of CDR-Agent may be further improved with larger models in the future (potentially with stronger capabilities). 

\begin{table*}[t]
\caption{Comparison of various LLM choices for CDR-Agent in CDR execution.
}\label{tab:model_choice}
\small
\begin{center}
\begin{tabular}{|l|l|l|l|l|l|l|}
\hline
\multirow{2}{*}{} & \multicolumn{3}{c|}{Synthetic Data} & \multicolumn{3}{c|}{CDR-Bench}\\
\cline{2-7}
& GPT-4o & GPT-4o-mini & GPT-4 & GPT-4o & GPT-4o-mini & GPT-4\\
\hline
Sensitivity & 0.983 & 0.987 & 0.966 & 0.683& 0.717& 0.786\\
\hline
Specificity & 0.687 & 0.583 & 0.711 & 0.983& 0.967& 1.0\\
\hline
$T_{\text{exe}}$ & 0.78 & 1.29 & 3.05 & 0.64& 0.75& 1.89\\
\hline
\end{tabular}
\end{center}
\end{table*}

\begin{figure*}[t]
    \centering
    \begin{minipage}[b]{0.44\textwidth}
        \centering
        \includegraphics[width=\textwidth]{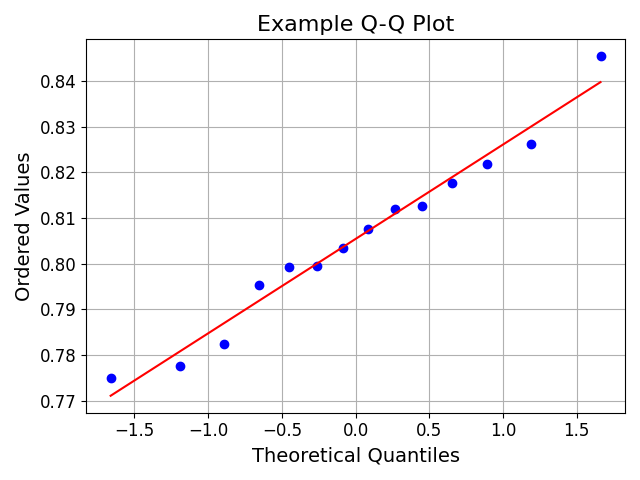}
    \end{minipage}%
    ~ 
    \begin{minipage}[b]{0.44\textwidth}
        \centering
        \includegraphics[width=\textwidth]{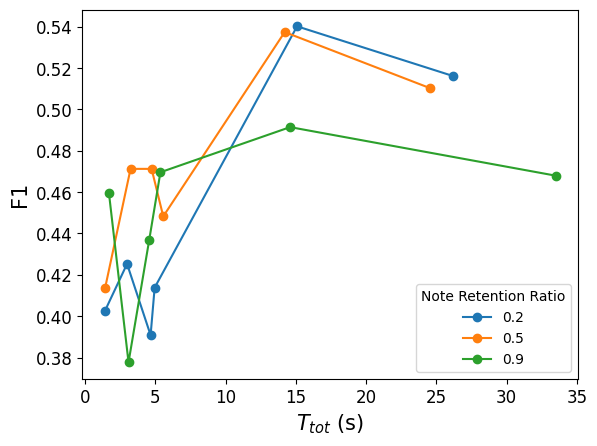}
    \end{minipage}
    \caption{(Left) An example Q-Q plot demonstrating that a Gaussian distribution is a reasonable choice for modeling similarity scores of irrelevant CDRs.
    (Right) Trade-off between F1-score and computation time on a held-out set of CDR-Bench for varying numbers of random sampling iterations and note retention ratios.}
    \label{fig:design_choice}
\end{figure*}

\paragraph{Design Choices and Further Improvements}
\label{subsec:design_choices_and_improvement}
In the CDR selection step, we model the similarity scores of irrelevant CDRs using a Gaussian distribution.
To validate this choice, we analyze randomly sampled clinical notes by reviewing the Q-Q plots of similarity scores for all CDRs deemed irrelevant to each note.
As illustrated by the example Q-Q plot on the left of Figure \ref{fig:design_choice}, the similarity scores align closely with the normal reference line (shown in red), showing that the Gaussian distribution is a reasonable choice.
However, accurate estimation of the mean and variance of the Gaussian distribution requires a sufficient number of similarity scores.
To improve the robustness of CDR selection, we apply repeated random truncations to clinical notes when computing similarity measures.
This approach not only increases the number of samples (though dependent)
for more reliable estimation of the distribution parameters, but may also reduce redundancy in the notes arising from verbose documentation.
As shown on the right of Figure \ref{fig:design_choice}, we vary the number of random sampling iterations across [2,5,8,10,30,50] while testing different note retention ratios.
This experiment is conducted on a held-out set consisting of 20\% of clinical notes randomly sampled from CDR-Bench, using the same CDR-Agent settings as in our main experiments.
Across all retention ratios, increasing the number of iterations generally improves CDR selection performance, albeit at the cost of longer computation times.
Notably, when focusing on smaller segments of notes (lower note retention ratio), performance improves more sharply as the number of iterations increases.
These results suggest that, for real-world CDRs with noisy or verbose documentation, applying a greater number of random truncations, particularly with shorter note segments, can lead to more robust CDR selection, should the time budget allow.

\paragraph{Incorporation of Additional Knowledge}
To account for the variability in real-world clinical notes, where medical terms often have alternative phrasings and abbreviations, we explore enhancing CDR-Agent with additional knowledge. Specifically, we augment the textual descriptions of CDRs by appending a list of synonymous terms for key indicators, prefixed with ``Keywords to consider often include:''. For example, for distal radial fracture, we include alternatives such as DRF, fx distal radius, distal radius fracture, and for wrist swelling, we add swollen wrist, wrist puffiness, enlarged wrist. These keyword expansions were generated using GPT-4o with a one-shot example and later verified by a clinician. Incorporating this additional knowledge improved exact match accuracy by 4\% and F1 score by 2\% on CDR-Bench. This demonstrates CDR-Agent’s flexibility in integrating domain-specific enhancements, making it more robust to linguistic variability in clinical text.

\section{Related Work}\label{sec:related_work}

In addition to the related work mentioned in the introduction, there are several other studies also highlight the advancements in clinical decision support systems (CDSS) using LLMs.
For example, LLMs enhanced with clinical guidelines show improved performance in tasks like COVID-19 outpatient treatment \cite{oniani2022enhancing, Wang2024}.
For another example, LLM-generated explanations increase doctors' agreement rates with the diagnoses \cite{Umerenkov2023}.
Moreover, various studies employ retrieval augmented generation (RAG) with LLMs or integrated non-knowledge-based models like fine-tuned BERT for medical predictions \cite{Ong2024, LEVRA2025113}.
Differing from these studies, our research specifically focuses on a unique clinical decision-making scenario, trauma-related CDR selection and execution in emergency rooms, where both the accuracy and efficiency of the decision-making process are critical.

\section{Conclusion}

In this chapter, we introduced CDR-Agent, an LLM-based system designed to autonomously select and execute Clinical Decision Rules (CDRs) based on clinical notes. To address the lack of public benchmarks for validating automated CDR selection and execution, we developed two high-quality datasets, laying the groundwork for future research in LLM-driven clinical decision support. Using these datasets, we demonstrated that CDR-Agent not only selects relevant CDRs efficiently, but makes cautious yet effective imaging decisions by minimizing unnecessary interventions while successfully identifying most positively diagnosed cases, outperforming traditional LLM prompting approaches.

While our framework is generalizable across clinical scenarios, we chose to focus on trauma care as a compelling case study due to its multidisciplinary nature and the availability of multiple, organ-specific CDRs.
By bridging AI capabilities with clinical expertise, CDR-Agent has the potential to support decision-making not only in specialized trauma centers but also in diverse clinical settings where expertise may be limited.
This work represents an important step toward deploying safe, transparent, and context-aware AI tools in frontline multidisciplinary trauma care.

\section*{Acknowledgements}
We gratefully acknowledge partial support from NSF grant DMS-2413265, NSF grant DMS 2209975, NSF grant 2023505 on Collaborative Research: Foundations of Data Science Institute (FODSI), the NSF and the Simons Foundation for the Collaboration on the Theoretical Foundations of Deep Learning through awards DMS-2031883 and 814639, NSF grant MC2378 to the Institute for Artificial CyberThreat Intelligence and OperatioN (ACTION), NSF grant 1910100, NSF grant 2046726, NSF AI Institute ACTION No. IIS-2229876, NIH (DMS/NIGMS) grant R01GM152718, AI Safety Fund, and a Berkeley Deep Drive (BDD) Grant from BAIR and a Dean’s fund from CoE, at UC Berkeley.
AZ additionally acknowledges support from NSF RTG Grant 1745640.
Dr. Aaron Kornblith is a consultant and co-founder for Capture Dx and supported by the Eunice Kennedy Shriver National Institute of Child Health and Human Development of the National Institutes of Health under award number K23HD110716 (AK). This information or content and conclusions are those of the author and should not be construed as the official position or policy of, nor should any endorsements be inferred by HRSA, HHS or the U.S. Government.

\bibliography{ref}
\bibliographystyle{icml2025}


\end{document}